# Lane Change Intention Recognition and Vehicle Status Prediction for Autonomous Vehicles

RentengYuan, Mohamed Abdel-Aty, *Member, IEEE,* Xin Gu, Ou Zheng, *Member, IEEE,* and Qiaojun Xiang

*Abstract*— Accurately detecting and predicting lane change (LC)processes of human-driven vehicles can help autonomous vehicles better understand their surrounding environment, recognize potential safety hazards, and improve traffic safety. This paper focuses on LC processes, first developing a temporal convolutional network with an attention mechanism (TCN-ATM) model to recognize LC intention. Considering the intrinsic relationship among output variables, the Multi-task Learning (MTL)framework is employed to simultaneously predict multiple LC vehicle status indicators. Furthermore, a unified modeling framework for LC intention recognition and driving status prediction (LC-IR-SP) is developed. The results indicate that the classification accuracy of LC intention was improved from 96.14% to 98.20% when incorporating the attention mechanism into the TCN model. For LC vehicle status prediction issues, three multi-tasking learning models are constructed based on MTL framework. The results indicate that the MTL-LSTM model outperforms the MTL-TCN and MTL-TCN-ATM models. Compared to the corresponding single-task model, the MTL-LSTM model demonstrates an average decrease of 26.04% in MAE and 25.19% in RMSE.

## I. INTRODUCTION

It can be expected that, for an extended period of time, vehicles with varying levels of automation will coexist on the roads [1, 2]. During the transition period, assisting intelligent driving vehicles to understand and predict changes in the behavior of human-driven vehicles is particularly critical for driving decisions. LC is a common driving behavior that leads to two-dimensional spatial (longitudinal and lateral) interaction between vehicles. The LC process consists of a series of continuous, complex maneuvering actions that significantly impact road traffic efficiency and safety [3, 4]. Accurately identifying and predicting lane change processes can help intelligent driving vehicles anticipate potential safety risks and execute appropriate response strategies.

The LC behavior is a time-varying, continuous maneuvering process [5, 6]. LC intention recognition has been a challenging problem in traffic engineering since it is hard to observe directly. There are two types of information that are typically utilized to identify LC intentions: vehicle dynamics indicators and driver characteristic indicators. Vehicle dynamics indicators include steering wheel angle, steering velocity, lateral velocity, turn signal, and brake pedal position [7-10]. In addition to the difficulty of obtaining certain information directly from human-driven vehicles (e.g., steering wheel angle, steering velocity, etc.), the reliability of the data obtained is also difficult to guarantee. For example, turn signal usage is reported to be between 44% and 40% in the US and China, respectively[8,11]. The driver characteristic indicators consist of head movement, eye movement, body gestures, and even electroencephalography[12-19]. Such information can only be gathered through sensors or driving simulation experiments. Inevitably, experimental settings constrain these investigations, such as potential concerns with low data quality, high cost, and small sample size, making it difficult to generalize and apply the research findings.

With the advancement of technology, traffic system monitors, and road users can obtain massive, real-time, individualized, and high-precision vehicle trajectory data. Lane change trajectory prediction has attracted a lot of attention over the past few years[20-24]. However, vehicle status indicators are more frequently utilized than vehicle trajectory information in practical engineering applications, such as real-time risk assessment, driving decisions, and vehicle control[25-29]. The research on driving status prediction can be classified into two categories: speed prediction[30-33], and steering angle prediction[34-36]. Previous modeling frameworks have required separate prediction models for each metric to predict the driving status, resulting in significant training time and potential conflicts between the prediction results of different metrics. In fact, these driving status indicators are interrelated, especially for vehicles that are performing lane changing behavior [37, 38]. Considering the correlation among indicators, developing a multi-task prediction model to predict multiple indicators simultaneously is necessary to reduce model training time and improve prediction performance [39].

To our knowledge, no study has been conducted specifically to focus on LC vehicle status indicator prediction. In this

Manuscript received 15 June 2023; revised ** ** 2023; accepted ** 2023. Date of publication **** 2023; date of current version ** ** 2023. This work was supported in part by the National Natural Science Foundation of China (71871059), the Shanxi Provincial Innovation Center Project for Digital Road Design Technology (202104010911019), the Postgraduate Research & Practice innovation Program of Jiangsu Province (No.KYCX22_0270), and the China Scholarship Council (CSC). *(Corresponding author: Qiaojun Xiang).*

RentengYuan, and Qiaojun Xiang are with the Jiangsu Key Laboratory of Urban ITS, School of Transportation, Southeast University, Nanjing, 210000, P. R. China. (e-mail: rtengyuan123@126.com; xqj@seu.edu.cn). RentengYuan currently is a visiting scholar in the University of Central Florida.

Mohamed Abdel-Aty and Ou Zheng are with the Department of Civil, Environmental and Construction Engineering, University of Central Florida, 12800 Pegasus Dr #211, Orlando, FL 32816, USA (e-mail: M.aty@ucf.edu; ouzheng1993@knights.ucf.edu).

Xin Gu is with the Beijing Key Laboratory of Traffic Engineering, Beijing University of Technology,Beijing, 100124,China(e-mail: guxin@bjut.edu.cn).

Color versions of one or more of the figures in this article are available online at http://ieeexplore.ieee.org

paper, the vehicle status was characterized using six variables, including the longitudinal velocity($v_x$), lateral velocity($v_y$), longitudinal acceleration($a_x$), lateral acceleration ($a_y$), vehicle heading($\theta$), and yawRate($\triangle\theta$). This paper using vehicle trajectory data aims to build a unified approach to LC intention recognition (LC-IR) and LC vehicle status prediction (LC-SP). The contribution of this paper is threefold.

●Firstly, a new unified modeling framework for Lane Change Intention Recognition and Status Prediction (LC-IR - SP) based on vehicle trajectory data is proposed. A new vehicle trajectory dataset (*CitySim Dataset*) is employed to develop the LC-IR-SP model. As far as we know, this is the first study to combine lane change intention recognition and status prediction.

●Secondly, to effectively capture crucial temporal features, this study integrates the attention mechanism into TCN networks, resulting in the development of a novel TCN-ATM model specifically designed for LC (Lane Change) intention recognition. The incorporation of the attention mechanism enhances the model's capacity to selectively focus on and extract pertinent temporal information.

●Thirdly, considering the inherent interdependencies among outcome variables, this study constructs three multi-task learning models (MTL-LSTM, MTL-TCN, MTL-TCN-ATM) for predicting driving status variables. To our knowledge, no studies simultaneously considered the intrinsic relationship between outcome factors to predict driving status indicators.

The rest of this paper is structured as follows. Section II presents a brief literature review. The data collection and post-processing are described in Section III. In Section IV, a new unified modeling framework for Lane Change Intention Recognition and Status Prediction is proposed. The experimental results and discussion are included in Section V, Section VI draws out the conclusions of this study.

II. RELATED RESEARCH

There are three kinds of methods, including dynamic or kinematic models [36-42], statistical models [7,10,15, 43-45], and machine learning methods [46-52], that have been widely used for LC intention recognition. The method based on dynamic or kinematic models detects the vehicle's motion by considering the kinematic relationship among parameters (e.g. position, velocity, acceleration), and the different forces (the longitudinal and lateral tire forces, or the road banking angle) that affect the vehicle motions. As classical statistical methods, multinomial logit models and Bayesian theory are utilized to predict the lane change probability. To capture the inherent characteristics of time series, three machine learning methods, including Hidden Markov Model (HMM), Support Vector Machines (SVM), and Long Short-Term Memory (LSTM), have been widely used. The commonly used models and their performance are listed in Table I.

TABLE I
A summary of the representative research for LC intention recognition

| Study | Data | Method | Number of Samples | Advance time | Accuracy (%) |
|---|---|---|---|---|---|
| [40] | Image | CNN | 637 | -- | 73.97 |
| [41] | Image | GoogleNet & LSTM | 714 | 3.76s | 74.46 |
| [42] | | Vision-cloud | --2(Pts) | -- | 79.2 |
| [43] | Simulator | AT-BiLSTM | --2(5Pts) | 3s | 93.33 |
| [10] | | BN | --(1 Pt) | -- | 95.4 |
| [44] | | HMM | --(58 Pts) | -- | 83.22 |
| [7] | | SVM | 139(6 Pts) | 1.3 | 80 |
| [12] | Naturalistic | EBiLSTM | 201(3 Pts) | 0.5 s | 96.1 |
| [8] | | HMM | 642(50 Pts) | 0.5s | 90.3 |
| [45] | | LSTM | 814(6 Pts) | -- | 88.26 |
| [46] | | RVM | 903(8 Pts) | 3s | 88.51 |
| [47] | | NN | Above 1000 | -- | 73.33 |
| [48] | | LSTM | --- | 2.5s. | 92.40 |
| [49] | | Logit | Above 1000 | -- | 66.41 |
| [50] | Trajectory | LSTM | Above 1000 | 2s | 86.21 |
| [51] | | HMM | 3410 | 6s | 94.4 |
| [5] | | Extra trees classifier | Above 1000 | 2s | 82 |
| [21] | | SVM | 351 | 3s | 85 |

Notes: --represents Not reported; Pt represent the participants

The summary of the literature in Table I reveals several valuable conclusions. First, vision-based LC behavior recognition methods exhibit lower classification accuracy than other methods. Second, because simulators and natural experiments are limited by the small number of experimental participants and high data homogeneity, ensuring the model's generalizability is challenging. Third, machine learning-based models have better classification accuracy compared to statistics-based models. Fourth, the Long Short-Term Memory (LSTM) Networks are widely used for lane change intention recognition and have made great progress in improving the accuracy of the LC behavior recognition but still have excellent potential to improve classification accuracy.

The LSTM approach has two limitations: the gradient vanishing problem and the inability to perform parallel computation [52]. To overcome the above two constraints, Temporal Convolutional Networks (TCNs), first proposed by [53], have attracted considerable interest. TCNs are designed for processing sequential data, such as time series or natural language[54-56]. With dilated causal convolution layers, TCNs effectively capture long-term dependencies over multiple time scales in the input sequences. Following that, TCNs have achieved significant promotion in both regression and classification tasks, involving forecasting carbon prices[56], predicting wind speed[57, 58], and diagnosing power converter faults[59].

Although vehicle status indicators can be extracted from the predicted trajectories, the method is restricted by error accumulation and lags in trajectory prediction results. Changes in vehicle velocity and driving direction will cause changes in vehicle trajectory. Extracting status indicators from predicted vehicle trajectories necessitates an extended prediction time of vehicle trajectories. Research reported that minor positioning errors might significantly affect extraction indicators [60]. Hence, building independent prediction models for driving status indicators is essential to improve predictive performance. The dilemma encountered by traditional end-to-end models is that to predict multiple indicators, a given model may be repeatedly trained to predict different indicators with the same input parameters, leading to computational redundancy and higher costs. To address this issue, Multi-task Learning (MTL) model was proposed first by Rich Caruana (1997) and involved training a model to learn multiple related tasks simultaneously. As a promising area in machine learning, MTL aims to improve the performance of multiple related learning tasks by leveraging useful information among them[61]. The tasks can be supervised, semi-supervised, or unsupervised and the model is designed to take advantage of shared representations across the various tasks in order to improve performance. [62]employed the MTL framework for traffic prediction, achieving up to 18% and 30% improvement in short- and long-term predictions.[63],[64] and [65] demonstrate the effectiveness of the MTL structure in travel time prediction areas. However, the ascendancy of MTL in driving status has not been tested. To fill the gap, this paper considers the MTL framework for LC-driving status prediction.

## III. NATURALISTIC DRIVING DATA

The publicly available *CitySim* dataset [66] is used in this research. The *CitySim* dataset is a drone-based vehicle trajectory dataset extracted from 12 locations with a sampling frequency of 30 Hz. With six lanes in two directions, a sub-dataset Freeway-B collected in Asia[66,67] is chosen to verify the performance of our proposed model. A snapshot of the freeway-B segment is shown in Figure 1.

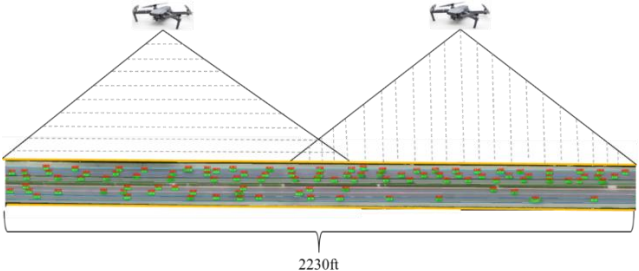

Fig. 1. A snapshot of the freeway-B segment

The freeway-B dataset is collected simultaneously with two UAVs over a 2230-ft basic freeway segment. Totally 5623 vehicle trajectories are extracted from 60 minutes of drone videos. This study focuses on lane change processes. A total number of 1023 vehicle trajectories are extracted ultimately from the freeway-B dataset, including 545 lane-change (LC) vehicle trajectories (240 left lane change (LLC)vehicle trajectories and 305 right lane change (RLC) vehicle trajectories) and 478 lane-keeping (LK) vehicle trajectories. Lane-keeping vehicle trajectories are randomly extracted.

*A. Data Processing*

Four significant steps are further employed for data processing with extracted vehicle trajectory data.

1) Removing abnormal data. The freeway-B dataset is collected from two stitched drone videos. The vehicle trajectory with the difference of adjacent frames greater than one is removed to avoid the effects of frame misalignment or skipping.

2) Data smoothing. Minor positioning errors might significantly affect extraction indicators [60]. To reduce the negative effect of errors, a moving average (MA) method is used to smooth the trajectory, and the moving average filter is set to 0.5s. A comparison of the original trajectory and processed trajectory is shown in Figure 2.

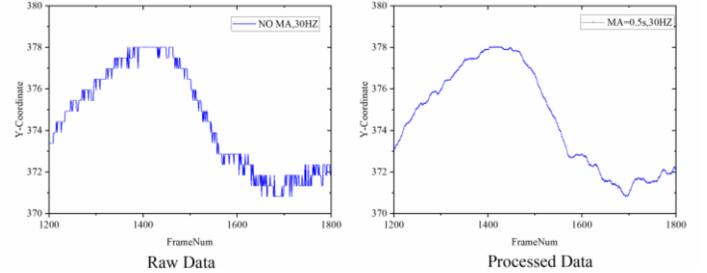

Fig. 2. Comparison of original trajectory and processed trajectory

3) Indicator calculation. To accurately describe the vehicle driving status, six indicators are extracted from the two-dimensional (i.e. longitudinal and lateral) vehicle position coordinates, including the longitudinal velocity ($v_x$), lateral velocity ($v_y$), longitudinal acceleration($a_x$), lateral acceleration ($a_y$), vehicle heading (θ), and yawRate(△θ). Furthermore, a non-linear low-pass filter is employed to reduce the negative effect of measurement errors [68]. First, the vehicle speed at the *t-th* frame is calculated and can be formulated as.

$$v_n(t) = \frac{s(t+n) - s(t-n)}{2 \cdot nT} \quad (1)$$

Where *t* represents the current frame, T is a constant, representing 1/30s in this research, *n* represents the time-step; $s(t-n)$ represents the vehicle's position in the frame $t-n$, where *n* takes different values, a vector $\{v_1(t), v_2(t), ..., v_N(t)\}$ (In this paper, *n* is set to 8)will be obtained. Thus, the vehicle velocity $v(t)$ at the *t-th* frame is calculated by taking the median of all N time steps. The lateral velocity ($v_y$) and longitudinal velocity ($v_x$) can be determined based on the change in the lateral and longitudinal positions of the vehicle, respectively. With the calculated velocity, acceleration can be obtained as.

$$a(t) = \frac{v(t+1) - v(t-1)}{2 \cdot T} \quad (2)$$

The lateral acceleration ($a_y$) and longitudinal acceleration ($a_x$) also can be determined based on the change of $v_y$ and $v_x$, respectively. In addition, the vehicle heading can be calculated as,

$$\theta_n(t) = \arctan\left(\frac{y_H(t+n) - y_R(t-n)}{x_H(t+n) - x_R(t-n)}\right) \quad (3)$$

Where $\theta_n(t)$ represents the vehicle heading at the frame $t$, $y_H(t+n)$ is the vehicle head point longitudinal position in the frame $t+n$, $x_R(t+n)$ denotes vehicle tail point horizontal position in frame $t+n$. yawRate is used to represent the rate of change of the vehicle's steering wheel angle[25]. It is calculated as,

$$\Delta\theta(t) = \frac{\theta(t+1) - \theta(t-1)}{2T} \quad (4)$$

5) Normalization. Variations in magnitude and units among different metrics can have an impact on the outcomes of data analysis. To mitigate this issue, it is necessary to standardize all the indicators.

$$x' = \frac{x - min(x)}{max(x) - min(x)} \quad (5)$$

*B Input indicator*

The vehicle status is influenced by other vehicles in the driving environment [69]. To fully consider the impact of various factors, the input of the combined model consists of three parts: ego vehicle (E-vehicle) information, surrounding vehicle information, and relative position information. Surrounding vehicles include the closest preceding and following vehicles in the adjacent and the current lanes. The ego vehicle is the human-driven vehicle. This research aims to detect and predict the human-driven vehicle LC process. The six indicators ($v_x$, $v_y$, $a_x$, $a_y$, $\theta$, $\Delta\theta$) were calculated for each vehicle. Limited by the video coverage, some trajectory fragments of surrounding vehicles were not recorded. A categorical variable (0 means it has recorded trajectory information; 1 means the trajectory information is missing) is added to each surrounding vehicle indicating this phenomenon. For instance, when the ego vehicle first appeared, the following vehicle (F-vehicle) was not yet in the drone videos. The following vehicle status variable(*F-val*) is set to 1. Relative position information(*dw*) is the headway distance between the E-vehicle and other vehicles, as shown in Figure 3. If the corresponding vehicle is not recorded in drone video, the corresponding *dw* is set to 0. Ultimately, a total of 54 indicators are taken as input variables. More details can be obtained from Table II.

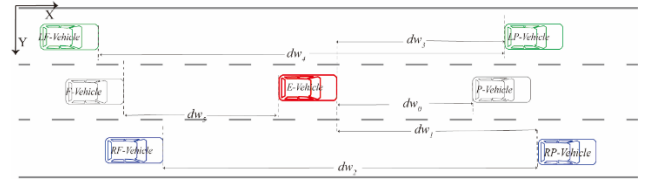

Fig.3 The headway distance between the E-vehicle and surrounding vehicles

TABLE II
Input indicators of the model

| Inputs Variable | Variable descriptions |
|---|---|
| E-, P-, F-, LP-, LF-, RP-, RF-$v_x$ | The longitudinal velocity of E-vehicle and surrounding vehicle (ft/ sec) |
| E-, P-, F-, LP-, LF-, RP-, RF-$v_y$ | The lateral velocity of E-vehicle and surrounding vehicle (ft/ sec) |
| E-, P-, F-, LP-, LF-, RP-, RF-$a_x$ | The longitudinal acceleration of E-vehicle and surrounding vehicle (ft/ sec $^2$) |
| E-, P-, F-, LP-, LF-, RP-, RF-$a_y$ | The lateral acceleration of E-vehicle and surrounding vehicle (ft/ sec $^2$) |
| E-, P-, F-, LP-, LF -, RP-, RF-$\theta$ | The heading of E-vehicle and surrounding vehicle (degree) |
| E-, P-, F-, LP-, LF -, RP-, RF-$\Delta\theta$ | The yawRate of E-vehicle and surrounding vehicle (degrees/sec) |
| $dw_0$, $dw_1$, $dw_2$, $dw_3$, $dw_4$, $dw_5$ | Space headway between E-vehicle and surrounding vehicle (ft) |
| P-, F-, LP-, LF -, RP-, RF-val | 0 means it has recorded trajectory information; 1 means the trajectory information is missing |

Note: "E-" represents the ego vehicle; "P-" represents the closest preceding vehicle in the same lane; "F-" represents the closest following vehicle in the same lane; "LP-" represents the closest preceding vehicle in the adjacent left lane; "LF-" represents the closest following vehicle in the adjacent left lane; "RP-" represents the closest preceding vehicle in the adjacent right lane; "RF-" represents the closest following vehicle in the adjacent right lane;

## IV. MATH

In this section, this paper first proposes a new modeling framework for Lane Change Intention Recognition and vehicle Status Prediction (LC-IR -SP). Then four commonly used time series classification methods are introduced, respectively. Furthermore, by incorporating an attention mechanism, a new novel TCN-ATM (TCN with attention mechanisms) model is proposed. Based MTL framework, several multitask learning models, MTL-TCN-ATM, MTL-TCN, and MTL-LSTM, are developed to predict LC vehicle status. Finally, the commonly used evaluation metrics are presented.

*A Modeling framework*

Figure 4 presents the framework of the proposed Lane Change Intention Recognition and Status Prediction (LC-IR -SP) model, which consists of two core modules: the Lane Change Intention Recognition (LC-IR) module and the Lane Change Status Prediction (LC-SP) module. The LC-IR module is a classification model used to recognize whether the vehicle produces LLC intention or RLC intention. When the LC-IR module detects that a vehicle generates a lane change intention, the LC-SP module will predict the LC vehicle driving status. The LC-SP module consists of separate multi-task learning and single-task learning models for sequence-to-sequence prediction. Multi-task learning models are employed to predict related variables. Unrelated variables were predicted separately using a single-task model.

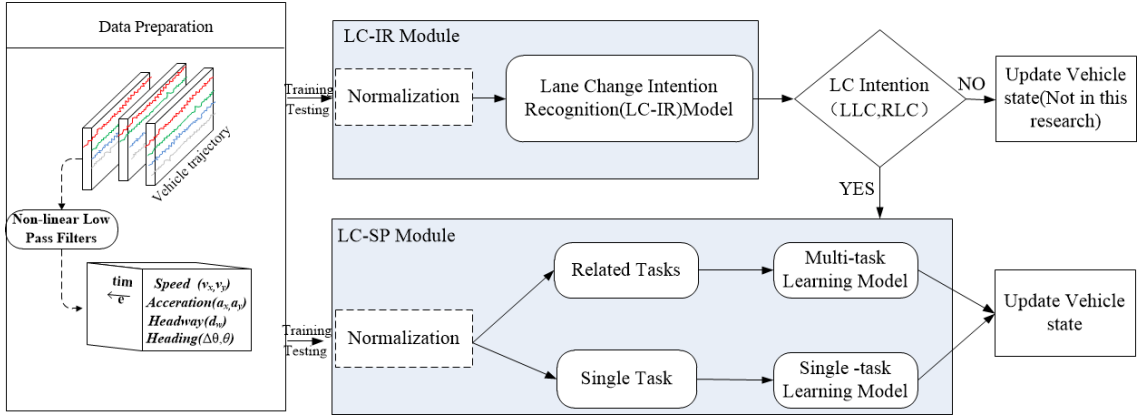

Fig.4. Modeling framework based on deep learning

1) LC-IR module. LC intention is divided into three categories: lane keeping (LK), left lane changing (LLC), and right lane changing (RLC). The generation of LC intention is a complex process, and it is influenced by other vehicles in the driving environment[15, 43]. As mentioned above, there are 54 indicators used as input variables. Lane Change Intention Recognition can be conceptualized as a classification issue based on multivariate time-series data. The function of the LC-IR module is defined as,

$$L_t = \varphi\left(\ell \cdot S_{t-\Delta t:t}\right) \tag{6}$$

The output $L_t$ represents the LC intention of the ego vehicle at time t, which is labeled as 1, 2, and 3 for LK, RLC, and LLC; $S_{t-\Delta t:t} = \{v_x, v_y, \cdots, \Delta\theta\}_{t-\Delta t:t}$ represents the input variables described in Section III, the notation $t-\Delta t:t$ denotes a time-series of the indicator from time $t-\Delta t$ to time t; $\ell$ denotes the parameter vector; $\varphi()$ represents the mapping relationships.

2) LC-SP module. The driving status is represented by six indicators: $v_x, v_y, a_x, a_y, \theta$, and $\Delta\theta$. Velocity ($v_x, v_y$) and heading ($\theta$) can be regarded as macroscopic indicators that reflect the aggregated effects of prior driving behavior. Meanwhile, acceleration ($a_x, a_y$) and yawRate ($\Delta\theta$), used as microscopic indicators, indicate the driving behavior that the driver is about to perform, reflecting changes in the throttle, brake pedal, and steering angle of the vehicle, respectively. Predicting LC vehicle driving status requires the simultaneous prediction of these six indicators ($v_x, v_y, a_x, a_y, \theta$, and $\Delta\theta$).

The LC process usually lasts for several seconds. With a 1s interval (indicators take an average of 60 frames), lane-change vehicle status in the next 2s is predicted in this study. For instance, taking the longitudinal velocity ($v_x$) and the lateral velocity ($v_y$) as an example, the function of the LC-SP module is defined as,

$$\begin{aligned}\left(v_{y,t+1}, v_{y,t+2}\right) &= g_1\left(\xi_1 \cdot R_{t-\Delta t:t}\right) \\ \left(v_{x,t+1}, v_{x,t+2}\right) &= g_2\left(\xi_2 \cdot R_{t-\Delta t:t}\right)\end{aligned} \tag{7}$$

Where $v_{x,t+1}, v_{x,t+2}$ represent the longitudinal speed of the ego vehicle at time t+1 and time t+2, respectively; $v_{y,t+1}, v_{y,t+2}$ represent the lateral speed of the ego vehicle at time t+1 and time t+2, respectively. Compared with (6), the input $R_{t-\Delta t:t}$ has an additional variable $L(t-\Delta t:t)$, which denotes the LC intention from time t-Δt to time t. $g_1()$ and $g_2()$ represent the mapping relationships. $\xi_1$ and $\xi_2$ denote the parameter vector. The expected six output variables ($v_x, v_y, a_x, a_y, \theta, \Delta\theta$) are simultaneously influenced by the same surrounding environment. By intelligently leveraging the inherent relationships between variables, it becomes possible to enhance prediction accuracy effectively.

*B Classification model*

LC intention recognition is a multivariate time series classification problem. The indicators that require classification exhibit high dimensionality. Selecting the appropriate model for this particular issue in machine learning applications can be a complex and challenging task. There are three main methods commonly used: Support vector machines (SVM), Long Short-Term Memory Networks (LSTM), and Temporal Convolutional Networks (TCN).

1) Support vector machines

Support vector machine (SVM) is a supervised machine learning algorithm that is primarily used for classification tasks[7, 21]. The main idea of SVM is to find an optimal hyperplane by mapping vectors to a higher-dimensional space. The hyperplane could effectively separate data points of different classes, and on either side of this separating hyperplane, two parallel hyperplanes are established.

2) LSTM Methods

LSTM adopts a gating mechanism that selectively retains or forgets information, effectively enhancing the long-term dependency modeling capability of traditional RNNs. It could be employed individually to address such sequence-to-sequence prediction and time series classification issues. A typical LSTM block is configured mainly by an input gate $i_t$, forget gate $f_t$ and output gate $o_t$. These gates are computed as follows,

$$i_t = \sigma(W_i x_t + U_i h_{t-1} + b_i) \quad (8)$$

$$f_t = \sigma(W_f x_t + U_f h_{t-1} + b_f) \quad (9)$$

$$o_t = s(W_o x_t + U_o h_{t-1} + b_o) \quad (10)$$

Where $\sigma$ represents the sigmoid activation function; $x_t$ represents the input sequence at time $t$; $h_{t-1}$ represents the hidden state; $W$ is the parameter matrix at time $t$ and represents the input weight; U is the parameter matrix at time $t$-1 and represents the recurrent weight; $b_i, b_f$, and $b_o$ represent bias. The internal update state of the LSTM recurrent cells can be expressed as:

$$c_t = f_t \odot c_{t-1} + i_t \odot \tilde{c}_t \quad (11)$$

$$h_t = o_t \odot \tanh(c_t) \quad (12)$$

Where $\odot$ represents vector element-wise product, $c_t$ is the memory cell at time $t$-1, $\tilde{c}_t$ is the candidate memory at time t, $h_t$ is the outcome at time t.

3) Temporal convolutional networks

TCN consists of causal convolution and dilated convolution[53,54]. Causal convolutions are used to ensure the temporal dependencies of the input data. Given an input sequence $x_0, x_1, \ldots, x_n$ and the corresponding output sequence $y_0, y_1, \ldots, y_n$, the causality constraint causal convolutions ensure that the output $y_t$ at time t is only determined by the input sequence $x_0, x_1, \ldots, x_t$. The one-dimensional fully-convolutional network (1DFCN) architecture is employed to produce the same length output as the input [70]. The TCNs can be expressed as,

$$TCNs = 1D\ FCN + causal\ convolutions \quad (13)$$

Using causal convolution, it is theoretically possible to generate TCNs. However, the receptive field of causal convolution is constrained, making it difficult to capture the correlation between points in a long-term temporal sequence. Hence, dilated convolutions were added to causal convolutions, enabling an exponentially large receptive field. For a filter $f: \{1,2,\ldots,k-1\}$, the dilated convolution operation F on the element $s$ of a 1-D sequence $x \in R^n$ is formulated as,

$$F(s) = (x *_d f)(s) = \sum_{i=0}^{k-1} f(k) \cdot x_{s-d \cdot i} \quad (14)$$

Where $d$ is the dilation parameter and is used to control the size of the interval, $k$ is the filter size and represents the number of convolution kernels, * is the convolution operator, $s - d \cdot i$ accounts for the direction of the past. A dilated convolution will be backward to a full convolution when $d = 1$. The dilated causal convolution structure is depicted in Figure 5.

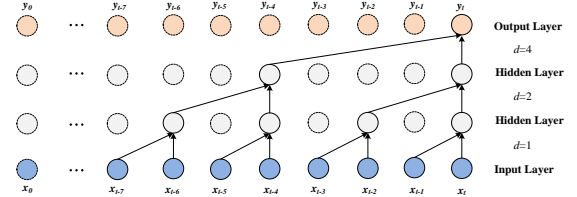

Fig.5. A dilated causal convolution with dilation factors d = 1,2,4 and kernel size k = 2 [71]

As shown in Figure 5, the kernel size is set to 2, and the depth of the causal convolution is 3. The convolution indicated that the output at time t is associated with the input points from time t-7 to time t. Residual blocks are used to address disappearance and gradient expansion in TCNs. Utilizing techniques such as longer convolutional kernels and residual connections allows TCN to capture long-term dependencies. As shown in Figure 6, the rectified linear unit (ReLU) is utilized as an activation function, and batch normalization is used as the convolutional filter. A 1x1 convolution is added in the residual block when the input and output data have different lengths.

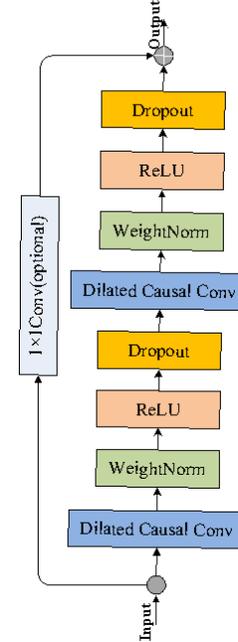

Fig.6. TCN residual block

By adjusting dilation parameters, the amount of information received by the TCN can be changed. The receptive field of the TCN can be calculated as,

$$R_{field} = 1 + (K-1) \times N_{stack} \times \sum_i d_i \quad (15)$$

Where $R_{field}$ represents the receptive field of the TCN, K is the filter size, $N_{stack}$ represents the number of stacks, $d_i$ represents the dilation parameter in the $i$th layer.

*C TCN with attention mechanism model*

To prioritize important input elements and enhance model performance and generalization, this study introduces an attention mechanism into the TCN network, creating a novel TCN with attention mechanism (TCN-ATM) model. The attention mechanism can be understood as a straightforward weighted summation operation. The relevant equations are formulated as,

$$u_t = tanh(\omega * h_t + b) \quad (16)$$

$$a_t = softmax(u_t) \quad (17)$$

$$c = \sum_{t=1}^{n} a_t * h_t \quad (18)$$

Where $h_t$ represents the extracted features by TCN layers at time t, $\omega$ is the parameter matrix at time *t*, $a_t$ is the weight of $h_t$ and could be calibrated based on the impact of each data feature on the output. *c* denotes the weighted sum of features $h_t$ at time t. The structure of TCN-ATM is depicted in Figure 7.

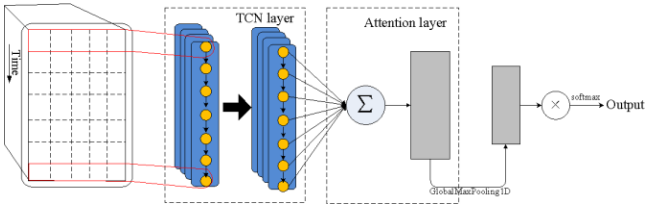

Fig. 7. The structure of the TCN-ATM model

The TCN-ATM model architecture consists of an input layer followed by a TCN layer and an attention layer. The TCN layer processes the input data while preserving the sequence information. An attention mechanism is then applied to the TCN output, capturing important features. Next, a global max pooling layer condenses the information into a fixed-length representation. Finally, a dense layer with softmax activation is added to produce class probabilities. This model architecture combines TCN and attention to effectively extract temporal patterns and make accurate predictions and classification tasks.

*D Multi-task Prediction Model*

MTL can be viewed as a generalization of multi-label learning and multi-output regression [61] and has the advantages of improving data efficiency, generalization ability, regularization ability, and overall performance[62, 72, 73]. It is designed to leverage a shared representation at the bottom layer and simultaneously enables learning multiple related tasks. For the *k*th task, the output $y_k$ in MTL can be expressed as:

$$y_k = h^k(f(x)) \quad (17)$$

Where the *f* function represents the shared-bottom network, $h^k$ denotes the *k*th tower network, and *x* is the input variable vector. Compared to single-task learning, multi-task learning framework can share information among different tasks, reduce training time, and improve the efficiency of data utilization [74]. Given three learning tasks, the comparison of single-task and multi-task learning architecture based on single-layer networks is shown in Figure 8. Tasks 1, 2, and 3 are three related output variables that share common input indicators. Using a single-task learning model requires building three separate models, but in the multi-task learning framework, only one model is required to be constructed with three outputs.

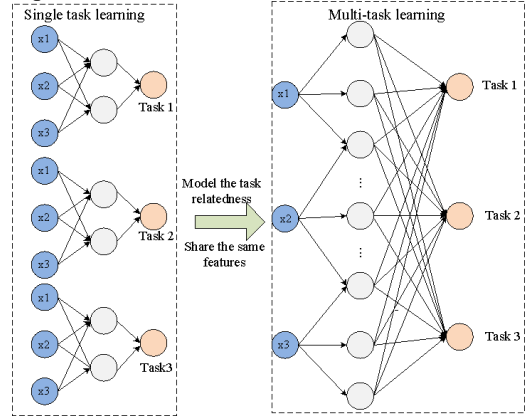

Fig.8. The comparison of single-task and multi-task learning architecture

The loss function is a critical component of multi-task learning. How to design the loss function for multi-task learning is crucial to determining the performance of the model. A common approach is to use a linear function to directly combine these loss functions, as shown,

$$Loss_{total} = \sum_i \omega_i Loss_i \quad (18)$$

Where $Loss_{total}$ represents the cumulative loss of all tasks, $Loss_i$ is the loss of the *i-th* task, and $\omega_i$ is the weight of the *i-th* task. By adjusting $\omega_i$, the model performance for the *i-th* task can be changed. For instance, if there is a main task in all the tasks, increasing the loss weight of the main task can improve the model performance for it. In this study, all tasks are considered equally important with assigned equal weights.

Based on the multi-task model framework, three multi-task models (MTL-LSTM, MTL-TCN, MTL-TCN-ATM) are first developed in this study. The multi-task model consists of a feature processing layer and a fully connected layer. The feature processing layer is used to extract temporal features, which can be LSTM or TCN or TCN-ATM. The general model structure is shown in Figure 9. In the LSTM layer, TCN layer, or TCN-ATM layer, multiple tasks benefit from shared parameters and features. This sharing allows the model to leverage common patterns and relationships across different tasks, leading to improved performance and efficiency. Then, fully connected layers are employed to output prediction results for each variable.

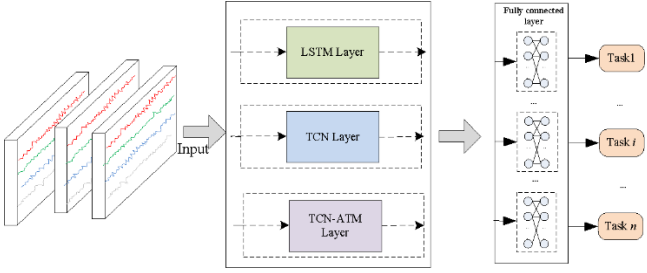

Fig.9. The general model structure of the three multi-task learning models

*E. Evaluation indexes*

The modeling framework proposed includes classification models and sequence prediction models. The performance of classification models is evaluated from two aspects. One is the overall performance of the classification, and the other is the recognition performance of each class [75]. The two indexes, precision and recall, are used to evaluate the detection performance of each class. The accuracy index measures the overall performance of the model. The three indexes can be calculated as follow,

$$Accuracy = \frac{T}{T + F} \quad (19)$$

$$Precision = \frac{TP}{TP + FP} \quad (20)$$

$$Rcall = \frac{TP}{TP + FN} \quad (17)$$

Where T represents the number of correctly classified samples, F represents the number of incorrectly classified samples, TP is the number of correctly classified samples in a given class, FP is the number of incorrectly classified samples in a given class, FN denotes the number of incorrectly classified samples in a given class. The two indexes, Mean Absolute Error (MAE) and Root Mean Square Error (RMSE) are employed to evaluate the performance of sequence prediction models. The definitions are as follows,

$$MAE = \frac{1}{N}\sum_{i=1}^{N}|y_i - \hat{y}_i| \quad (18)$$

$$RMSE = \sqrt{\frac{1}{N}\sum_{i=1}^{N}(y_i - \hat{y}_i)^2} \quad (19)$$

Where $y_i$ is the observed value of the *i*-th output, N is the number of outputs, $\hat{y}_i$ represents the predicted value of $y_i$. The prediction model with lower MSE and RMSE values performs better.

## V. RESULTS

To testify the feasibility of our modeling framework, the lane change intention recognition (LC-IR) model and the lane change status prediction (LC-SP) model are developed in this section, respectively. And we selected 545 LC vehicle trajectories and 478 LK vehicle trajectories for training and testing the lane change intention recognition model. However, only LC vehicle trajectories were used to train and test the LC vehicle status prediction model.

*A. Lane Change Intention Recognition*

The vehicle's lane change intentions are defined as some LC operational behavior produced before the lane change. To select an appropriate algorithm for classifying lane change intentions, this section compares the performance of four algorithms: LSTM, SVM, TCN, and TCN-ATM.

1) Lane-change intention labeling. In this study, the start time of the LC process is determined as the moment when the front boundary point of vehicles touches the lane boundary [27]. The annotation procedure determines the LC intention start time as 3 seconds forward the start time of LC processes. The start point of LC processes is considered to be the end point of LC intention. A total of 24,092 frames are labeled as RLC points, while 19,792 frames are labeled as LLC points. Figure 10 shows the detailed labeling processes. If the extracted sequence endpoint is located between the start time of the LC process and the LC intention start time (at least one RLC point or LLC point is included in the sequence), it is labeled as either LLC or RLC; Otherwise, it is labeled as LK.

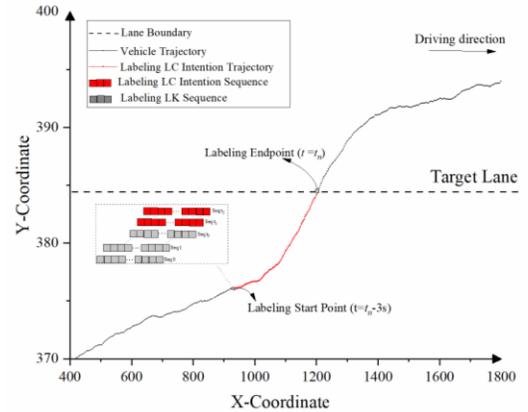

Fig.10. Lane-change intention labeling process

2) Results of LC intention recognition models. The dataset is randomly split into a training dataset and a test dataset with a ratio of 8:2. For training the LC intention classification model, eighty percent of total data is applied, and twenty percent of samples are used for testing the classification performance. The parameter setting will affect the performance of the model. To obtain optimal parameter settings, some sensitivity experiments are performed on four models, using the control variable method. The parameters are selected based on the metrics of classification accuracy and training time. The final model used should minimize the training time of the model (reduce the complexity of the model) without compromising the accuracy of the model. With an equal number of samples, all experiments are conducted using the same device. As an example, the impact of the number of epochs was evaluated with maintaining the same input durations (input time duration = 5s). Figure 11 depicted the results of the experiment. It is evident that when the number of epochs is set to a value greater than 50, the loss function does not exhibit significant changes. Hence, the epoch is set to 50. Finally, the kernel size is set to 2, the Batch size is set to 128, the Loss function is set to *categorical_crossentropy*, and the Adam optimizer was employed. The rate used in the dropout layer is 0.3. The size of dilated convolution interval is set to $\{2^1, 2^2, 2^3, \ldots, 2^n\}$, which depends on the input time series length. The number of filters is set to 64. The number of stacks of residual blocks is set to 1.

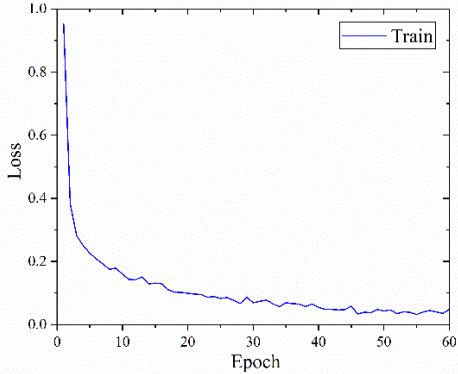

Fig.11. loss function

To investigate the effect of input sequence length on classification outcomes, the performance of LSTM, SVM, TCN, and TCN-ATM models with varying input durations and same parameter setting, is evaluated. With an interval of 15 frames, a total of 12 input lengths are extracted from 30 frames(1s) to 180 frames(6s).

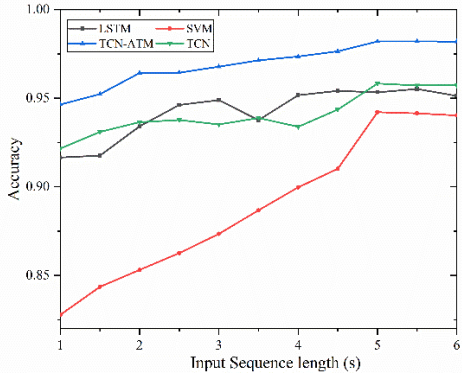

Fig.12. Accuracy comparison of LSTM, SVM, TCN, and TCN-ATM

Figure 12 illustrates the overall accuracy comparison results of the four models. It can observe that each model has good classification performance (above 80%), even though the four models are slightly different among different durations. With the same input data time scale, the TCN-ATM model outperforms TCN, SVM and LSTM models. The best classification accuracy was achieved for three models (LSTM, TCN, and TCN-ATM) when the input length was five seconds. Despite not attaining optimal accuracy using a 5-second input time length, the SVM algorithm exhibited marginal enhancements in classification accuracy. Hence, a time duration T = 150 frames (5s) was chosen as input sequence lengths. Finally, 22160 RLC sequences and 15410 LLC sequences were extracted. To maintain data balance, 18000 LK sequences are randomly extracted from the raw dataset. Using the training dataset, the ten-fold cross-validated method is employed for model training and evaluation.

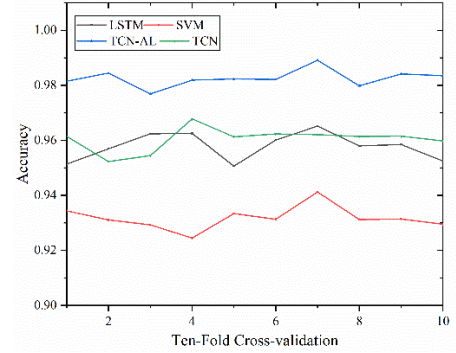

Fig.13. Ten-fold Cross-validation for LSTM, SVM, TCN, and TCN-ATM

Figure 13 illustrates the outcomes of a Ten-fold cross-validation analysis conducted on LSTM, SVM, TCN, and TCN-ATM models. The average accuracy for LSTM and SVM is 0.9568 and 0.9317, with standard deviations of 0.007 and 0.004, respectively. TCN and TCN-ATM algorithms have average accuracies of 0.9604 and 0.9825, with a standard deviation of 0.00001. The result indicated that TCN and TCN-ATM algorithms outperform LSTM and SVM and demonstrate more stability. With an input length of 150 frames, Figure 14 illustrates the confusion matrix for the four models using the validation set.

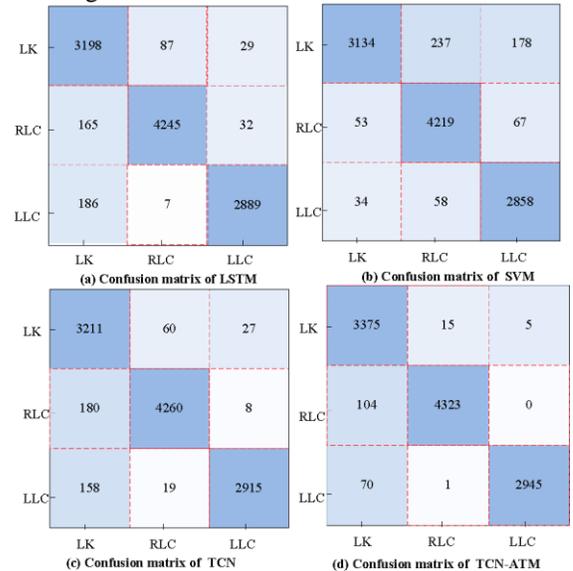

Fig.14. Confusion matrix of classification models

Errors in classifying LC intentions can be categorized into three categories: the misidentification of LK as LC (Type I), the misclassification of LC as LK (Type II), and the misidentification of LLC and RLC from each other (Type III). Figure 12 shows that the proposed TCN-ATM algorithm reduces the impact of Type II and Type III errors compared to LSTM, TCN, and SVM models. Type I errors have a significant impact on the accuracy of all four models. This error could originate from two sources. One is that the model correctly identifies the behavior of a failed lane change. The other could be attributed to the variations in individual lane change behaviors among drivers[76,77]. The LK process is influenced by factors such as driving style and driving ability, which can exceed the cognitive capabilities of the model, resulting in misjudgment. To provide a comprehensive assessment of classification performance, in addition to accuracy, other evaluation metrics such as precision, recall, and training time were evaluated through the confusion matrixes. The comparison results are displayed in Table III.

TABLE III
Evaluation results of LSTM, SVM, TCN, and TCN-ATM models

| Model | Type | Precision | Recall | Accuracy |
|---|---|---|---|---|
| LSTM | LK | 90.10% | 96.21% | 95.33% |
|  | RLC | 97.83% | 95.78% |  |
|  | LLC | 97.79% | 93.73% |  |
| SVM | LK | 88.31% | 97.29% | 94.21% |
|  | RLC | 97.23% | 93.46% |  |
|  | LLC | 96.88% | 92.10% |  |
| TCN | LK | 89.88% | 98.33% | 96.14% |
|  | RLC | 98.17% | 94.73% |  |
|  | LLC | 98.81% | 96.04% |  |
| TCN-ATM | LK | 95.09% | 99.41% | 98.20% |
|  | RLC | 99.63% | 97.65% |  |
|  | LLC | 99.83% | 97.64% |  |

Table III demonstrates that the overall performance of SVM, LSTM, and TCN is 94.21%, 95.33%, and 96.14%, respectively. The TCN-ATM model achieves an overall classification performance of 98.20%, exhibiting improvements of 2.06%, 2.83%, and 3.99% compared to the TCN, LSTM, and SVM models, respectively. On the other hand, when considered individually, the maximum deviation in classification precision is 7.73% for LSTM, 8.92% for SVM, 8.29% for TCN, and 4.74% for TCN-ATM. The maximum difference in recall index for each model is 2.48% for LSTM, 5.19% for SVM, 3.6% for TCN, and 1.87% for TCN-ATM. The result indicates that the TCN-ATM model provides more balanced results than other models. In summary, the proposed TCN-ATM model provides a promising solution for LC intention classification tasks, as it outperforms other models regarding classification accuracy.

### B. Lane Change Status Prediction

Lane change vehicle status involves six indicators: $v_x$, $v_y$, $a_x$, $a_y$, $\theta$, $\Delta\theta$. Predicting lane change status requires the simultaneous prediction of these six indicators. This section first uses the Pearson coefficient to investigate the relationship between those output indicators. Then three proposed multi-tasking learning models are used to capture the intrinsic relationship among these indicators.

1) Correlation Analysis. Multi-task learning (MTL) involves jointly learning multiple output indicators. The underlying assumption behind this approach is that all output indicators are related. Typically, the relationship between tasks could significantly affect the predictive quality of multi-task models [78]. Hence, the Pearson coefficient is employed to investigate whether there is an association between the variables. It can be expressed as,

$$r = \frac{\sum_{i=1}^{n}(x_i - \bar{x})(y_i - \bar{y})}{\sqrt{\sum_{i=1}^{n}(x_i - \bar{x})^2}\sqrt{\sum_{i=1}^{n}(y_i - \bar{y})^2}} \qquad (20)$$

Where $x_i$ represents the $i$th value of the indicator $x$, $\bar{x}$ and $\bar{y}$ represent the average value of indicator x and y, $r$ represents the Pearson coefficient and takes values in the range [-1,1]. The larger the absolute value of $r$, the stronger the correlation. In this research, only the indicators with an absolute value of Pearson coefficient greater than 0.2 are considered to be related[79, 80]. There are two types of vehicle information used to calculate the Pearson coefficient separately: LK vehicles and LC vehicles. The LC vehicle information used is the driving intention labeled segment defined in Fig.11. To mitigate the effect of sample imbalance on the results, 200 samples from each LK vehicle trajectory are extracted randomly.

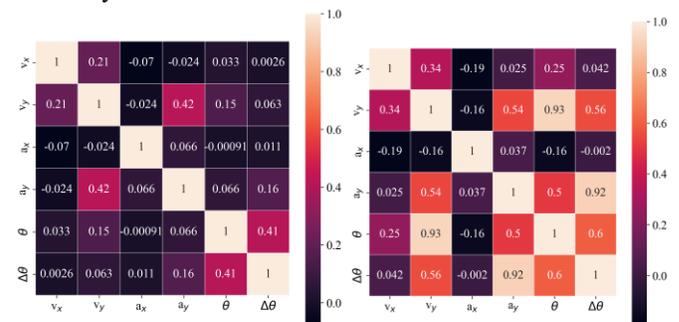

(a) Indicators Extracted from LK Sequences　　(b) Indicators Extracted from LC Sequences
Fig.15. Pearson coefficient heat map

The Pearson coefficient heat map is shown in Figure 15. It can be found that the Pearson coefficient between $v_x$ and $v_y$, $a_y$ and $v_y$, $\theta$ and $\Delta\theta$, which are greater than 0.2 in both types of sequences. The indicators extracted from the LK sequences exhibit a stronger correlation than those extracted from LC sequences. No significant correlation was found between lateral acceleration ($a_y$) and other indicators. Furthermore, the main discrepancies are observed between the heading-related indicators ($\theta$ and $\Delta\theta$) and the velocity-related indicators ($v_x$, $v_y$, and $a_y$). For lane-changing processes, strong correlations were observed between $v_x$ and $\theta$ (0.25), $v_y$ and $\Delta\theta$ (0.56), $a_y$ and $\theta$ (0.92), and $v_y$ and $\theta$ (0.93), indicating a close relationship between these variables. In contrast, no significant relationship was found in lane-keeping processes between heading-related indicators ($\theta$ and $\Delta\theta$) and velocity-related indicators ($v_x$, $v_y$, and $a_y$). The result could be explained by the fact that drivers have to adjust their driving direction and velocity during a lane change to achieve the desired purpose. However, during the lane-keeping phase without a specific task, the changes in heading and speed are random and separate from each other.

2) Results of LC status prediction models. With a focus on the purpose of the study, only sequences labeled as RLC and LLC are utilized in this section. Eighty percent of the extracted samples are used to train the model, while the remaining samples are used for validating the performance. The input sequence length is set to 150 frames(5s). With a 1s interval (indicators take an average of 60 frames), lane-change vehicle status in the next 2s is predicted. Several experiments are performed to obtain optimal parameter settings. The batch size is set to 64, and training epochs are set to 30. The loss function is mean_squared_error, and the optimizer is Adam. For TCN and TCN-ATM layer, the number of filters is specified as 64. For the LSTM layer, the number of neurons in the hidden layer is set to 64, and the depth of LSTM is set to 2.

The output indicators consist of six variables: $v_x$, $v_y$, $a_x$, $a_y$, $\theta$, $\Delta\theta$. Among these variables, $v_x$, $v_y$, $a_y$, $\theta$, and $\Delta\theta$ are related output variables, and $a_x$ is not correlated with other variables. In practice, three models, including LSTM, TCN, and TCN-ATM, could be utilized separately to address such sequence-to-sequence prediction issues. Given the expected output variables are simultaneously influenced by the same surrounding environment, three multi-task models (MTL-LSTM, MTL-TCN, MTL-TCN-ATM) are trained to improve prediction accuracy in this study. Three single-task learning models are used for comparison. The prediction results are listed in Table V.

TABLE V
Model result comparison

| Model | Metrics | Task |||||| 
|---|---|---|---|---|---|---|---|
| | | $v_x$ | $v_y$ | $a_x$ | $a_y$ | $\Delta\theta$ | $\theta$ |
| LSTM | MAE | 2.817 | 0.572 | 1.256 | 0.692 | 2.375 | 1.420 |
| | RMSE | 3.926 | 0.704 | 1.662 | 0.937 | 3.049 | 1.845 |
| MTL-LSTM | MAE | **1.288** | **0.502** | -- | 0.632 | 2.042 | **0.838** |
| | RMSE | **1.712** | **0.684** | -- | 0.866 | 2.510 | **1.080** |
| TCN | MAE | 2.977 | 0.596 | 1.064 | 0.669 | 8.159 | 1.894 |
| | RMSE | 4.134 | 0.793 | 1.402 | 0.922 | 10.60 | 2.408 |
| MTL-TCN | MAE | 1.982 | 0.547 | -- | 0.648 | 2.945 | 1.566 |
| | RMSE | 2.534 | 0.751 | -- | 0.916 | 3.875 | 2.062 |
| TCN-ATM | MAE | 1.749 | 0.561 | **0.975** | 0.875 | **0.560** | 1.002 |
| | RMSE | 2.080 | 0.693 | **1.235** | 1.188 | **0.799** | 1.293 |
| MTL-TCN-ATM | MAE | 17.18 | 1.751 | -- | 0.869 | 0.601 | 1.775 |
| | RMSE | 19.88 | 2.081 | -- | 1.183 | 0.858 | 1.464 |

From a single-task prediction perspective, the TCN-ATM model demonstrates superior performance compared to LSTM across various metrics ($v_x$, $v_y$, $a_x$, $\Delta\theta$, $\theta$). Additionally, it exhibits a significant reduction in Mean Absolute Error (MAE) and Root Mean Squared Error (RMSE) when compared to the TCN model. Notably, the TCN-ATM model excels in predicting the longitudinal acceleration $a_x$, with an RMSE value of 1.235 ft/s². These results highlight the effectiveness of incorporating the attention mechanism into the TCN model, thereby enhancing its performance in single-task learning. Consequently, the TCN-ATM model can be considered a practical and reliable option for single-index forecasting tasks.

From a multi-task prediction perspective, the MTL-LSTM model outperforms the MTL-TCN and MTL-TCN-ATM models for indicators $v_x$, $v_y$, $a_y$, and $\theta$. The MTL-TCN-ATM model demonstrates optimal prediction results for the $\Delta\theta$ indicator, while performing poorly for other indicators. With an RMSE value of 2.510 degree/s and an MAE value of 2.042 degree/s, the MTL-LSTM model shows considerable space for improvement in terms of the $\Delta\theta$ indicator. To compare the performance of single-task learning models with multi-task learning models, the improvement ratio is defined as follows:

$$p_i = 1 - \frac{m_i}{s_i} \quad (21)$$

Where $m_i$ represents the evaluation index (RMSE, MAE) value of $i$-th task using multi-task model, $s_i$ represents the evaluation index (RMSE, MAE) value of $i$-th task using the corresponding single-task models, $p_i$ is evaluation index improvement ration of task $i$ using MTL model comparing to single-task model. A positive value of $p_i$ indicates that the MTL model outperforms the corresponding single-task model in predicting task i, while a negative value indicates the opposite. Table VI presents the result of the improvement in prediction performance.

TABLE VI
Performance improvement rate of prediction (%)

| Model | Index | $v_x$ | $v_y$ | $a_y$ | $\Delta\theta$ | $\theta$ |
|---|---|---|---|---|---|---|
| MTL-LSTM vs LSTM | MAE | 54.28 | 12.24 | 8.67 | 14.02 | 40.99 |
| | RSME | 56.39 | 2.84 | 7.58 | 17.68 | 41.46 |
| MTL-TCN vs TCN | MAE | 33.42 | 8.22 | 3.14 | 63.90 | 17.32 |
| | RSME | 38.70 | 5.30 | 0.65 | 63.44 | 14.37 |
| MTL-TCN-ATM vs TCN-ATM | MAE | -882.2 | -212.12 | 0.69 | -7.32 | -77.15 |
| | RSME | -855.77 | -200.29 | 0.42 | -7.38 | -13.23 |

As is evident in Table VI, the proposed MTL-LSTM, and MTL-TCN over five indicators provide markedly increased performance compared to the corresponding single-task model. Specifically, the MTL-LSTM model demonstrates an average decrease of 26.04% in MAE and 25.19% in RMSE, while the MTL-TCN model exhibits an average reduction of 25.2% in MAE and 24.49% in RMSE. The performance improvement resulting from considering the relationship between output variables may be critical to accurately predicting driving status. However, the performance of the MTL-TCN-ATM model is much lower than that of the TCN-ATM model. The decrease in performance resulting from introducing attention mechanisms in multi-task learning can be attributed to issues such as task competition and conflicts, optimization challenges, and feature conflicts[81]. Different tasks may require attention to different features or information. When incorporating attention mechanisms, it becomes necessary to address feature conflicts among tasks to ensure that the attention mechanism can properly focus on and capture the relevant features for each task. In addition, when attention becomes excessively focused on one task, the important features of other tasks may be neglected, resulting in performance degradation.

## VI. CONCLUSION

LC behavior is a fundamental driving operation that largely affects traffic efficiency and safety. Accurately detecting and predicting lane change (LC)processes of the surrounding vehicles can help autonomous vehicles better understand their surrounding environment, recognize potential safety hazards, and improve traffic safety. In this paper, the LC vehicle status was characterized using six variables, including the longitudinal velocity ($v_x$), lateral velocity ($v_y$), longitudinal acceleration($a_x$), lateral acceleration ($a_y$), vehicle heading ($\theta$), and yawRate ($\triangle\theta$). Using vehicle trajectory data, this paper developed a unified modeling framework for lane-change intention recognition (LC-IR) and lane-change status prediction (LC-SP). To accurately identify LC intention, a novel TCN-ATM model was first utilized in this research. Considering the intrinsic relationship between outcome factors, three MTL models (MTL-LSTM, MTL-TCN, and MTL-TCN-ATM) were constructed to predict LC vehicle status. A total number of 1023 vehicle trajectories was first extracted from the *CitySim* dataset to validate the reliability of the proposed model. The Pearson coefficient was conducted to investigate the relationship between the output variables. Both training time and classification accuracy were utilized as metrics to evaluate the performance of the model.

For the LC intention recognition issues, this study conducted a comprehensive comparison of SVM, LSTM, TCN, and TCN-ATM models. The Ten-fold cross-validated method was employed to ensure robustness in model training and evaluation. With an input length of 150 frames, the proposed TCN-ATM model achieves an impressive overall classification performance of 98.20%. Compared to the LSTM, SVM, and TCN models, the results demonstrate that the TCN-ATM model reduces the impact of Type I and Type III errors, demonstrating a higher accuracy. For the LC driving status prediction issue, six metrics are extracted from the vehicle trajectory to characterize the driving status in this paper. The Pearson coefficient was employed to investigate the relationship between six output indicators. The result indicated a close relationship between the heading-related indicators ($\theta$ and $\Delta\theta$) and the velocity-related indicators ($v_x$, $v_y$, and $a_y$). To capture the intrinsic relationship of output indicators, tthis research developed three multi-task models: MTL-LSTM, MTL-TCN, and MTL-TCN-ATM. The results showed that the proposed TCN-ATM models could be considered a practical and reliable option for single-index forecasting tasks. The MTL-LSTM model outperforms the MTL-TCN and MTL-TCN-ATM model for indicators $v_x$, $v_y$, $a_y$, and $\theta$. With an average reduction of 26.04% and 25.19% in the MAE and RMSE, respectively. The proposed MTL-LSTM over five indicators provides markedly increased performance compared to the corresponding single-task model.

The research shows that the novel TCN-ATM model outperforms LSTM, SVM, and TCN models in lane change intention recognition. Considering the correlation of related indicators could improve the prediction accuracy and training efficiency of the model. According to the obtained index $v_x$, $v_y$, $a_y$, $a_x$, $\theta$, and $\Delta\theta$, the real-time traffic conflict index can be calculated [3]. According to the index $a_y$, $a_x$, $\theta$, and $\Delta\theta$, it can be determined whether the driver has taken the avoidance behavior. The developed model holds great potential in enhancing autonomous vehicles' perception and prediction capabilities and improving vehicle control strategies. This study also has some study limitations. In the multi-task learning model, we use the same weights for the loss function of each task. To eliminate the effect of magnitude on the prediction results, all input and output vectors are normalized. In the future, the prediction accuracy can be further improved by using the adaptive loss function. For instance, if there is a main task in all the tasks, increasing the loss weight of the main task could improve the model performance.